\documentclass[twoside,11pt]{article}

\usepackage{blindtext}
\usepackage{graphicx}
\usepackage{amsmath}
\usepackage{amssymb}
\usepackage{booktabs}
\usepackage{multirow}
\usepackage{natbib}
\usepackage[normalem]{ulem}
\useunder{\uline}{\ul}{}

\usepackage[preprint]{jmlr2e}

\usepackage{lastpage}

\ShortHeadings{}{Scale-Translation Equivariant Network for Oceanic Internal Solitary Wave Localization}
\firstpageno{1}

\begin{document}

\title{Scale-Translation Equivariant Network for Oceanic Internal Solitary Wave Localization}

\author{\name Zhang Wan \email zwan004@e.ntu.edu.sg \\
       \addr University of Birmingham\\
       Nanyang Technological University
       \AND
       \name Shuo Wang \email s.wang.2@bham.ac.uk \\
       \addr University of Birmingham
       \AND
       \name Xudong Zhang \email zhangxd@qdio.ac.cn \\
       \addr Institute of Oceanology, Chinese Academy of Sciences
}


\maketitle

\begin{abstract}
Internal solitary waves (ISWs) are gravity waves that are often observed in the interior ocean rather than the surface. They hold significant importance due to their capacity to carry substantial energy, thus influence pollutant transport, oil platform operations, submarine navigation, etc. Researchers have studied ISWs through optical images, synthetic aperture radar (SAR) images, and altimeter data from remote sensing instruments. However, cloud cover in optical remote sensing images variably obscures ground information, leading to blurred or missing surface observations. As such, this paper aims at altimeter-based machine learning solutions to automatically locate ISWs. The challenges, however, lie in the following two aspects: 1) the altimeter data has low resolution, which requires a strong machine learner; 2) labeling data is extremely labor-intensive, leading to very limited data for training. In recent years, the grand progress of deep learning demonstrates strong learning capacity given abundant data. Besides, more recent studies on efficient learning and self-supervised learning laid solid foundations to tackle the aforementioned challenges. In this paper, we propose to inject prior knowledge to achieve a strong and efficient learner. Specifically, intrinsic patterns in altimetry data are efficiently captured using a scale-translation equivariant convolutional neural network (ST-ECNN). By considering inherent symmetries in neural network design, ST-ECNN achieves higher efficiency and better performance than baseline models. Furthermore, we also introduce prior knowledge from massive unsupervised data to enhance our solution using the SimCLR framework for pre-training. Our final solution achieves an overall better performance than baselines on our handcrafted altimetry dataset. Data and codes are available at \url{https://github.com/ZhangWan-byte/Internal_Solitary_Wave_Localization}.
\end{abstract}


\section{Introduction}\label{sec_1}

Oceanic Internal Solitary Waves (ISWs) are gravity waves beneath the ocean surface, produced by disturbance of fluid with stable density stratification \cite{peng2023analysis,osborne1980internal,apel1985sulu,alford2015formation,guo2014review}. Due to its generation mechanism, ISWs usually carry a high magnitude of energy and can have large amplitude up to 200 meters. As a result, artifacts and activities of human beings, such as offshore oil platforms, submarines, and marine fishery, can be greatly affected by the existence of ISWs. In addition, as a common phenomenon in nature, ISWs also play an important role in ocean dynamics including ocean energy budget, and coastal ecosystem function \cite{bib36}. Therefore, detecting and further locating ISWs is of great significance. Conventional methodologies for researching ISWs often involve laboratory experiments, in situ observations, theoretical and computational analyses, and remote sensing measurements \cite{peng2023analysis}. However, the recent surge in on-orbit satellites and sensors has propelled ocean remote sensing into the big-data era, empowering data-driven approaches to leverage this vast resource effectively \cite{bib7}. 

The past decade witnessed a rise and surge in deep learning (DL), along with its fast development in remote-sensing imagery. As such, researchers have been developing automated ISW recognition methods based on optical and Synthetic Aperture Radar (SAR) images \cite{bib2,bib3,bib7}, where ISW localization is formulated as common computer vision tasks such as object detection \cite{bib2}, semantic segmentation \cite{bib7}, edge detection \cite{bib3}, etc. Although demonstrated high effectiveness, imagery-based solutions are intrinsically fragile to cloud obstruction and limited by the data duration or coverage. The partial observation issue poses a great challenge to global and retrospective studies on ISWs. Besides, images only provide two-dimensional information as they merely look down from the top, which further restricts the performance. In light of this, recent studies started to focus on altimetry-based approaches \cite{bib8,bib9,bib10}. High-precision instruments on satellites can not only record visual signals but also detect other ground signals such as the physicochemical properties of the sea surface. As a result, it not only captures ISW signatures but also provides a third-dimensional surface information caused by ISWs \cite{bib9,bib10}, which provides a new route of locating ISWs automatically with data-driven methods. Despite typically being perceived as low-resolution and characterized by a low signal-to-noise ratio (SNR), the substantial accumulation of historical altimetry data proves advantageous in training machine learning (ML) models for the localization of ISWs. More importantly, altimetry is less susceptible to cloud obstruction, offering global coverage that facilitates comprehensive analyses of ISWs across the globe.
Therefore, we employ low-resolution global altimetry data for ISWs localization in this study. To the best of our knowledge, no work has detected or located ISWs using altimeter data due to the high cost of labeling and low spatial resolution. The challenges, however, mainly lie in limited labels as these supervised signals require intensive and laborious handcrafts. As a result, data efficiency and learning efficiency are utterly significant in our scenario. Furthermore, a strong learner is needed due to the low resolution in exchange for easy access.

Recent developments in considering symmetries in machine learning provide elegant, effective, and general principles for neural network design. In the seminal paper \cite{cohen2016group}, a convolutional neural network equivariant to the p4m group is proposed where group theory is introduced to the machine learning community. The core idea of group equivariance is considering inherent symmetries by forcing the learned mapping to be equivariant to certain transformation groups. For instance, the p4m group contains all elements that rotate an image for 90$^{\circ}$, so that a p4m equivariant neural network can map transformed inputs to outputs that also transform predictably (it's rotating arbitrarily times of 90$^{\circ}$ in this case). Later studies on group equivariance further extend this principle to more groups such as the roto-translation group \cite{fuchs2020se} and Euclidean group \cite{weiler2019general,satorras2021n}. By imposing symmetries, neural networks are given strong priors, which benefit sample efficiency and generalization capacity, since the plausible search space is profoundly decreased by introducing geometric constraints. Therefore, it takes fewer labeled data to learn and achieves better generalization in samples not encountered before.

In this study, we address the aforementioned challenges by leveraging symmetries as an inductive bias to devise an efficient equivariant convolutional neural network (ECNN) tailored for localizing internal solitary waves (ISWs) in altimetry data. We partition spatially contiguous observations into independent windows, formulating ISWs localization into an intra-window multi-class classification challenge. Highlighting the role of incorporating symmetries as an inductive bias, we aim to optimize data efficiency, learning efficiency, and generalization capacity. Specifically, we introduce scale and translation as pivotal symmetries within our problem formulation, leading to the scale-translation equivariant convolutional neural network (ST-ECNN). Two architectures of ST-ECNN are proposed for experiments and comparison. To enhance performance, we further integrate SimCLR \cite{bib27}, a self-supervised pre-training technique, to learn prior knowledge from massive unlabeled data. After meticulous and laborious labeling by expert knowledge, we propose an altimetry-based ISWs dataset which is employed to rigorously evaluate the proposed method with multiple rounds of k-fold cross-validation. The experimental findings demonstrate that our method consistently outperforms various baselines across selected metrics.

We believe our solution comprehensively addresses the common and primal challenge in remote sensing tasks: how to achieve good performance and generalization capacity with efficiency given a limited quantity of labeled data. Our main contributions are summarized as follows:

\begin{itemize}

    \item We propose a novel approach to detect and locate ISWs at the finest grain using altimetry instead of imagery, which allows comprehensive global and retrospective ISW analyses.
    \item A scale-translation equivariant convolutional neural network (ST-ECNN) is introduced to tackle the challenge of the limited quantity of supervised data by injecting symmetries into network design.
    \item We empirically show that SimCLR, a self-supervised pre-training technique employed to learn priors from massive unsupervised signals, can improve performance based on altimetry-based data and our task formulation. The pre-trained ST-ECNN achieves the best overall performance than baseline models.
\end{itemize}

The rest of this paper is organized as follows. In Section \ref{sec_2}, we review the existing studies on ISW localization. We also discuss related works on recently developed strong convolutional neural networks (CNN) for comparison. The employed techniques in this study, group equivariant neural networks, and contrastive pre-training, are also introduced. Our proposed method is discussed in Section \ref{sec_3}, including problem formulation and the symmetries inherent in our task, followed by the architecture design of ST-ECNN. We also discuss the evaluation problem under the imbalance settings in Section \ref{sec_3}, where we propose feasible measures as complementary metrics. Experiments are designed, conducted, and discussed in Section \ref{sec_4}. We finally summarize the conclusions and discuss future works in Section \ref{sec_5}.

\section{Related Work}\label{sec_2}

In this section, we first introduce the existing research progress on ISW localization and related machine learning methods for ISW-related tasks. Then, we discuss the state-of-the-art deep convolutional network architectures as baselines. The principle of inductive bias injection and theoretical modeling of symmetries using group theory is also discussed. A the end of this section, we introduce contrastive learning (CL) techniques to utilize massive unsupervised data as pre-training.

\subsection{Existing Methods for ISW Localization}\label{subsec_2_1}


Among the multiple approaches to studying the characteristics of ISWs, remote sensing has been widely focused due to its vast spatial coverage and abundant data accumulations \cite{peng2023analysis}. Previous ISW detection methods based on remote sensing extract and analyze ISW features from SAR images. In \cite{bib3}, Canny edge detection is used for ISW feature capture. Signal processing algorithms such as Fast Fourier Transform (FFT) and Continuous Wavelet Transform (CWT) are involved to characterize ISWs \cite{bib5}. Furthermore, nature-inspired algorithms have also been used for ISW detection. For instance, the particle swarm optimization algorithm is used with 2-D wavelet transform on Advanced Synthetic Aperture Radar (ASAR) data \cite{bib4}, which searches for the optimal solution indicating ISW locations. However, the aforementioned methods have become less focused due to the synergistic performance of the fast-developed deep neural networks (DNNs) and increasingly accumulated remote-sensing data \cite{bib7}. This new fast-evolving area brings stronger AI tools for image processing and starts to replace the traditional methods \cite{bib7}.

Computer vision has been a major beneficiary of the past decade's surge of deep learning. Image-based remote sensing techniques using deep learning therefore further enjoy advancements, which are then employed to detect ISWs in SAR and optical images. In \cite{wang2019fast}, ISW localization is formulated as an object detection task where the PCANet and SVM are combined to detect internal waves. In \cite{bib2}, ISW localization is further improved by Fast R-CNN, a prevalent object detection framework. A recent paper \cite{bib7} summarizes deep learning approaches for oceanic imagery, where two learning frameworks, U-Net and single-shot multi-box detector (SSD), are introduced for pixel-level classification (i.e. image segmentation) and object-level detection tasks. Both frameworks can annotate ISWs on images directly. The pixel-level classification aims to find all pixels that constitute ISWs, whereas the object-level detection returns the frame (or a bounding box) that contains ISWs in an image.

Although several deep learning algorithms have been applied for ISW localization, image-based approaches are highly affected by data quality and quantity. For example, the cloud obstruction of the sea surface can significantly limit the efficacy of ISW detection. In comparison with images, altimeter data are conventionally considered incapable of ISW detection due to their small periods and spatial scales \cite{bib9}. However, recent studies \cite{bib9,bib10, yu2022study} show that short-period ISWs can be detected by the synthetic aperture radar altimeter (SRAL) on board the Sentinel-3A. 
As such, a multi-modal neural network was proposed in \cite{bib8} for internal wave detection which considers both image and SAR altimetry data with higher along-track resolution. Nevertheless, it is interesting to find that the predictive model trained solely on altimeter data outperforms the combination of both sources \cite{bib8}. Inspired by this study, we propose to locate ISWs with altimetry signatures. However, the limited supervisory signals due to time-consuming handcrafts remain a primal challenge in our scenario. We shall introduce deep learning approaches and how to inject prior knowledge through symmetry for data efficiency, learning efficiency, and generalization capacity in the following sections.

Beyond ISW localization, deep learning has also been widely accepted as a novel solution in ISW-related problems, especially in inverse problems. In \cite{zhang2021machine}, a fully connected neural network (FCN) is employed to predict internal wave location based on satellite images. FCN is also used in \cite{zhang2020combination} to forecast propagation locations of internal waves from initial conditions. An ISW amplitude inversion model is constructed with shallow neural networks in \cite{zhang2022machine}. Despite directly learning from remote sensing data, the transfer learning technique is also incorporated to retrieve internal wave amplitude from satellite images by learning from simulation lab data. Although deep learning has been widely used in internal wave scenarios, the utilized methods are limited to FCNs in different scales since FCNs are suitable for formulations in the inversion problem. Furthermore, it is worth noting that existing solutions in ISW-related inverse problems can not conduct global historical retrospective analyses, because: 1) accumulated errors during iterative predictions which lead to inaccuracies of final results, and 2) huge computation expenditure due to the inefficient nature of propagation predictions (predicting the next location and status of ISWs based on current conditions).

Based on the previous studies in internal wave scenarios discussed above, it is observed that no recently developed deep learning methods are focused and employed. As such, we shall introduce several state-of-the-art convolutional neural networks as baselines in the next section.

\subsection{Deep Learning Architectures: ResNet and variants}\label{subsec_2_2}


With sufficiently large data, deep learning approaches often perform better than traditional ML algorithms, especially on unstructured data, such as images \cite{bib45,bib46,bib1}, text \cite{bib47,bib48,bib49}, videos \cite{bib50,bib51,bib52}, audios \cite{bib53,bib54} and graphs \cite{bib55}. Considering altimeter data is also unstructured, deep learning models can be competitive candidates to tackle challenges in ISW data. Besides, we further notice certain similarities between image data and altimeter data. As a consequence, we adopt Convolutional Neural Network (CNN) for feature extraction based on the following considerations: 1) Inter-regional information in data is important for ISW localization. Therefore, slicing original multi-variate sequence altimeter data into windows is necessary. It is impossible to infer the existence of an ISW at one location (i.e. using only one data point with several variables as features to predict ISW exists or not) without considering its neighborhood locations. Thus, similar to object detection in images, locating ISWs should consider a contiguous set of data points within a region, rather than a single data sample. 2) CNN can extract local features very well with convolution. The feature variables on the ISW data point are signally different than its neighbors, which is quite similar to the intensity variation of the edge on the image. The convolution operation is capable of detecting ``edge features", and can therefore reveal differences between target object and background signals. It may help detect ISWs within a region. Given these considerations, we transform the original sequence of ISW data into a set of windows of data in this paper. Each window containing altimeter data from multiple locations within the same region is treated as one input sample. More details about data preparation can be found in Section \ref{subsec_3_1}.


Deep Residual Neural Network (ResNet) and Bottleneck Transformer (BoTNet) are two significant developments on CNN. ResNet has demonstrated a superior feature capture capability and effectiveness by learning identity mapping with residual connection, leading to a deeper neural network \cite{bib1}. It has been widely used as the backbone network for various neural network architectures. Compared to VGG \cite{bib56} and GoogleNet \cite{bib57}, the residual connection enables ResNet to contain more layers by learning identity mappings, and thus provides better performance. Considering its superiority, we study ResNet as one of the potential solutions for ISW localization in this paper. 

Building on top of ResNet, BoTNet is a more recent attempt to introduce the idea of attention into CNN. In 2017, Transformer \cite{bib11} was first developed and has since become a widely used architecture in various fields such as computer vision (CV) and natural language processing (NLP). One significant component of the Transformer is its attention mechanism: Multi-Head Self-Attention (MHSA). In essence, the attention mechanism assigns weights to input data with self-awareness, which allows the model to pay more attention to important and local information. The newly proposed MHSA in \cite{bib11} uses a set of weighting matrices to linearly project original inputs into another set of spaces, then a dot-product attention mechanism is applied independently in each space. The results of each space are then concatenated to form the eventual output. Based on MHSA, the authors in \cite{bib12} proposed an attention-enhanced ResNet, i.e. BoTNet, where the convolution layers are replaced by global MHSA layers. Their experiments indicate a better performance of BoTNet than ResNet on image classification and segmentation tasks. With convolution and attention mechanisms, both local and global information for comparative differences in data can be well extracted and represented. Therefore, BoTNet is also considered as our potential solution in this work.

The development of ConvNeXt \cite{liu2022convnet} serves as a retrospective fight-back against Transformer. As one of the state-of-the-art architectures, ConvNeXt is purely constructed using convolution modules, where multiple engineering designs such as inverted bottleneck and large kernel size are included to improve performance. Due to its simplicity and impressive performance on benchmarks, we here consider ConvNeXt as a strong baseline for comparison.

\subsection{Symmetry as Inductive Bias}\label{subsec_2_3}

Inductive bias is the set of assumptions that the model utilizes to predict outputs of given inputs not encountered before \cite{mitchell1980need}. Incorporating prior knowledge as an inductive bias to restrict action space of neural networks has been proven beneficial in better representing underlying function, and thereby achieves higher learning efficiency, data efficiency, and generalization capacity \cite{bronstein2021geometric,roberts2022principles}. One great example is CNN whose inductive bias is that contiguous pixels are assumed to co-constitute low-level features including edges and dots. Such prior knowledge is explicitly included through convolutional filters, and the spatial locality leads to higher efficiency in learning as compared to MLPs, because feasible parameter searching space is profoundly decreased due to the mechanism it learns (i.e., learning a set of kernels for template matching). Given such a locality assumption broadly holds in the image domain, it levels up higher generalization capacity since incorrect parameter space has been disregarded.

From another point of view, CNNs are usually considered to respect translational invariance. Invariance refers to different inputs leading to the same output, while equivariance indicates that alterations in inputs result in predictable changes to outputs. However, as shown in \cite{biscione2021convolutional}, pure convolution operation obeys translational equivariance, while modern CNNs are not perfectly translational equivariant (due to sub-sampling operations such as pooling or striding) though they can learn to be. This further entails the importance of symmetry in the design of neural networks, since correct symmetry as inductive bias is significant for two important properties of neural networks: efficiency and generalization \cite{biscione2021convolutional}.

\begin{figure}[ht]%
\centering
\includegraphics[width=1.0\textwidth]{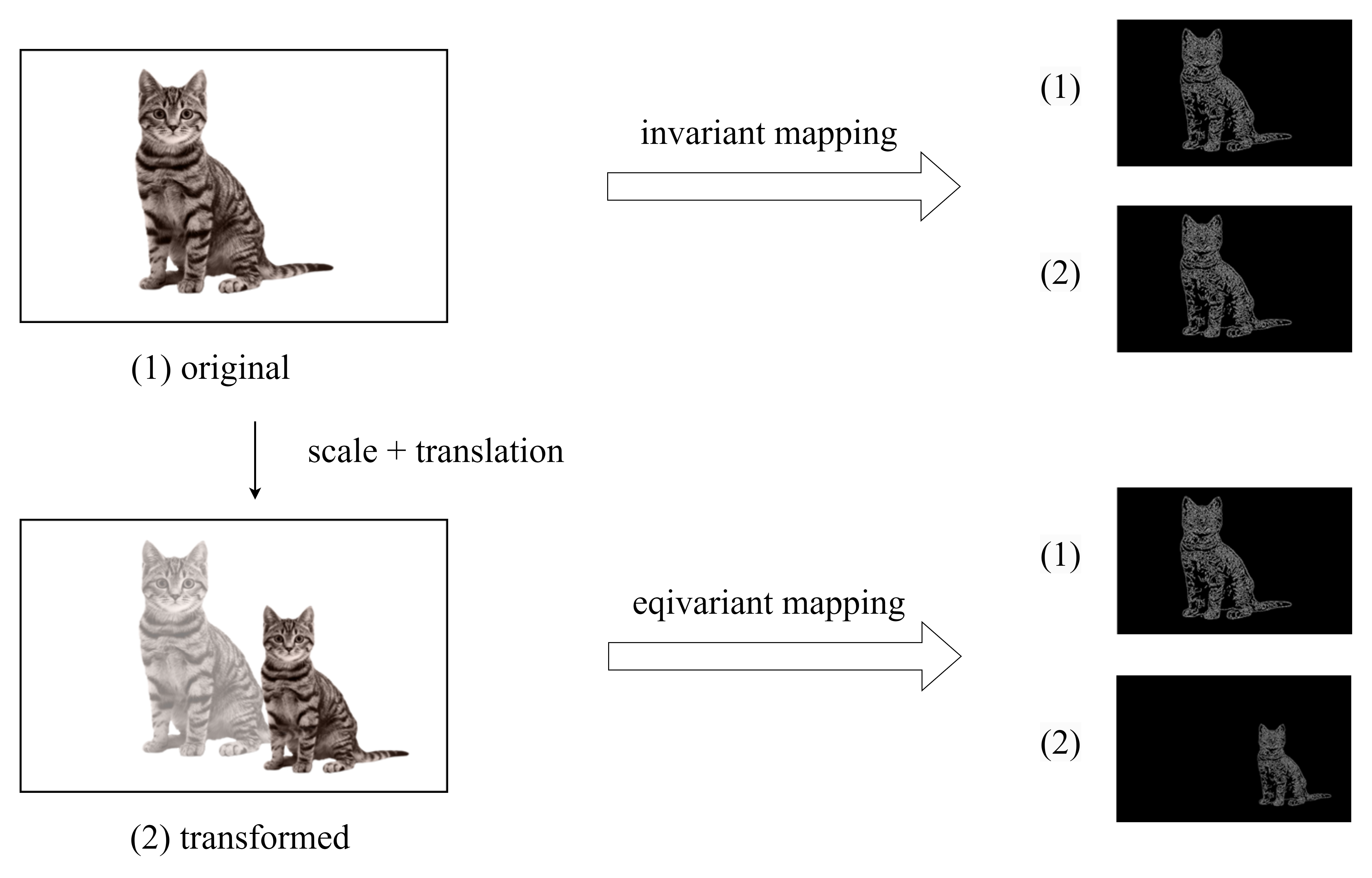}
\caption{\textbf{Equivariance and Invariance.} The left bottom image is obtained through scaling and translating the original image (1). \textbf{Invariance.} A transformed image leads to the same feature representation as that of the original image. \textbf{Equivariance.} Transformed image leads to a different while predictable feature, i.e., scaling and translating edge feature as equivalent to the transformation on inputs.}\label{fig:equi}
\end{figure}

Group equivariant convolutional networks (G-CNNs) \cite{cohen2016group} further extend the equivariance to the general symmetry groups. In \cite{cohen2016group}, group convolution is proposed to consider any operation (e.g. scale, translation, rotation) applied on inputs, which can be described by a certain group $\mathcal{G}$. In Fig. \ref{fig:equi}, we show an image of a cat being transformed using a combination of operations. A mapping is equivariant to a group if an input transformed by an element of the group corresponds to the output with an equivalent transformation, while invariance is a special case of equivariance where all transformations to input lead to equivalent output. We formally define the general form of group equivariance as follows: given a group $\mathcal{G}$ and its elements $g \in \mathcal{G}$ acting on set $\mathcal{X}$, a mapping $\phi: \mathcal{X} \rightarrow \mathcal{X}$ is equivariant to $\mathcal{G}$ if:

\begin{equation*}\label{eq:equi}
    \phi(gx) = g \phi(x), \forall x \in \mathcal{X}, \forall g \in \mathcal{G}.
\end{equation*}

Another way to illustrate equivariance is the commutativity of operation, as shown in Eq. \ref{eq:equi}. Similarly, we define invariance of $\phi$ to $\mathcal{G}$ if:

\begin{equation*}\label{eq:inv}
    \phi(gx) = \phi(x), \forall x \in \mathcal{X}, \forall g \in \mathcal{G}.
\end{equation*}

It is worth noting that invariant mappings remove all symmetries from the input, while equivariant mappings preserve symmetries. However, it is also noteworthy that the choice between equivariance and invariance is highly dependent on specific tasks. In this paper, we propose to employ a scale-translation equivariant convolutional neural network (ST-ECNN) to well tackle the challenge of efficient learning. Scale and translation are incorporated as symmetries because they are inherent to our data domain. We shall justify it and introduce our method in Section \ref{sec_3}.

\subsection{Contrastive Learning}\label{subsec_2_4}

Self-supervised learning is an area in unsupervised deep learning. It learns from supervised signals constructed by specifically designed tasks such as data reconstruction or complementing incomplete images, where no explicit label is involved. This may benefit our ISW task which has limited labeled data compared to the much larger size of unlabeled data. 

Contrastive learning (CL) is a key type of self-supervised learning, commonly used for pre-training \cite{bib60,bib61}. Pre-training refers to training a model based on a source task with a large amount of unlabeled data and using this model to help learn a target task with labeled data.  CL learns an encoder to project raw inputs without labels to a hypersphere representation space. In this hypersphere space, the distance between similar instances (positive pair) is smaller than that between dissimilar instances (negative pair). This is achieved by explicitly optimizing contrastive pair loss that measures the difference between positive pairs and negative pairs. By recognizing the similarities and differences between data samples, CL represents original data into high-level features without using labels. The encoder obtained from CL is subsequently used for learning the target task. In other words, the knowledge from unlabeled data is transferred to a DNN model by initializing its parameter weights with the weights in the encoder. 

There have been several CL frameworks proposed \cite{bib27,bib26,bib28,bib29}. Their mechanisms of training an encoder by optimizing a contrastive loss are similar, such as Noise Contrastive Estimation (NCE) \cite{bib41},:

\begin{equation}\label{eq_1}
    \mathcal{L} = \mathbb{E}_{x,x^{+},x^{-}}[-log(\frac{e^{f(x)^{T}f(x^{+})}}{e^{f(x)^{T}f(x^{+})}+e^{f(x)^{T}f(x^{-})}})]
\end{equation}

where $x^{+}$ and $x^{-}$ denote similar and dissimilar input data in $x$ respectively, and $f$ is the encoder. To optimize this loss function, the distance between positive instances (usually set to instances of the same class) is minimized while that of negative instances (usually set to instances of a different class) is maximized. To this end, the encoder we wish to have as mentioned above is achieved.

In addition to making better use of unlabeled data, researchers have also noticed the benefit of CL in solving class imbalance problems. For a highly imbalanced medical image learning task, CL was used to pre-train an encoder based on a re-sampled dataset and fine-tune the classifier \cite{bib30}. Another work employed CL on a typical class-imbalanced credit rating task and showed improved performance over previous methods \cite{bib31}. The latest study also reveals the robustness of self-supervised learning approaches to class imbalance \cite{bib37}. The advantages of CL shown in the existing literature motivate us to explore its effectiveness in ISW localization. 

In this paper, we use SimCLR \cite{bib27} to conduct CL pre-training. As a pervasive contrastive learning framework, it has been widely used for its simplicity and efficiency. SimCLR performs training in batches. Its objective is to train an encoder that maps original data into another discriminative space, in which the distance between any two new representations of samples is measured by cosine similarity. After data augmentation, all data samples are paired into positive pair-wise and negative pair-wise samples. Similarly to clustering, the training objective of an encoder in SimCLR is to minimize the normalized temperature-scaled cross-entropy loss \cite{bib29} (NT-Xent), i.e. minimizing the distance of positive pairs of samples and maximizing the distance of negative pairs of samples, as indicated in Eq. \ref{eq_2}. 

\begin{equation}\label{eq_2}
    \mathcal{L}_{i,j} = -log(\frac{exp(sim(z_i,z_j)/\tau)}{\sum_{k=1,k \neq i}^{2N}{exp(sim(z_i,z_k)/\tau)}})
\end{equation}

\noindent
where $z_i$ is the vector embedding projected by the encoder, and sim(·) is the cosine similarity function. The loss is computed for all positive pairs (i,j) and negative pairs (i, k) in one batch. In SimCLR, the positive pair contains the original data and its augmented form. The intuition behind this is that corrupted data still represents key properties for classification. The negative pairs, however, are constructed by sampling from this batch of data during training. Although it is at risk of selecting samples from the same class, the pre-training doesn't acquire accurate classification but only distinguishes between samples based on features. For vision tasks, it's rare to see identical cats, thus it's necessary to maintain substantial information for each data point even if they are in the same class. Generally, the larger the batch size is, the better encoder we can get, because a large batch constructs more pair-wise training samples.

$\tau$ in Eq. \ref{eq_2} is the temperature hyper-parameter inspired by Simulated Annealing (SA) for better optimization. Different from how it's used in simulated annealing, $\tau$ in NT-Xent is a pre-defined constant. A larger $\tau$ flattens the similarity curve more, namely, the distance between a positive pair of samples $z_i$ and $z_j$ will be curtailed more, leading to a mild distance distribution. The temperature coefficient $\tau$ should be carefully chosen. It determines the clustering degree of positive pairs. A large $\tau$ is not necessarily beneficial, because it may curtail the independence of samples.

\section{Methods}\label{sec_3}

In this study, we select altimetry-based data for ISW localization due to its accessibility (easy to obtain) and completeness (less affected by obscuring clouds). The challenge, however, mainly lies in the limited supervised signals due to laborious labeling. Considering the recent progress in efficient deep learning and self-supervised learning, we aim to tackle the challenges by injecting prior knowledge in twofold: symmetry incorporation and self-supervised pre-training.

In this section, we first describe the altimetry dataset used in this study. Based on the dataset, we formulate the learning task into a multi-class classification problem. A scale-translation equivariant convolutional network is proposed through symmetries inherent in our dataset.

\subsection{Problem Formulation}\label{subsec_3_1}

The dataset used in this work is collected from the ocean regions of the South China Sea, Andaman Sea, and Sulawesi Sea since large-scale ISWs are more frequently observed in these regions. Among all collected samples, 21.55\% contain ISWs, leading to a class imbalanced distribution. Each data sample contains the traditional altimeter information in Low-Resolution Mode (LRM) from the satellites Jason-2/3 due to its abundant historical data resource since 1992, while recently launched satellites with high-resolution instruments accumulate far less data than Jason-2/3. However, due to its limited resolution, the ISW signatures in LRM data are less clear than the recent SAR altimeter. We will need ML models with strong feature extraction capacity.

\begin{figure}[ht]%
\centering
\includegraphics[width=0.9\textwidth]{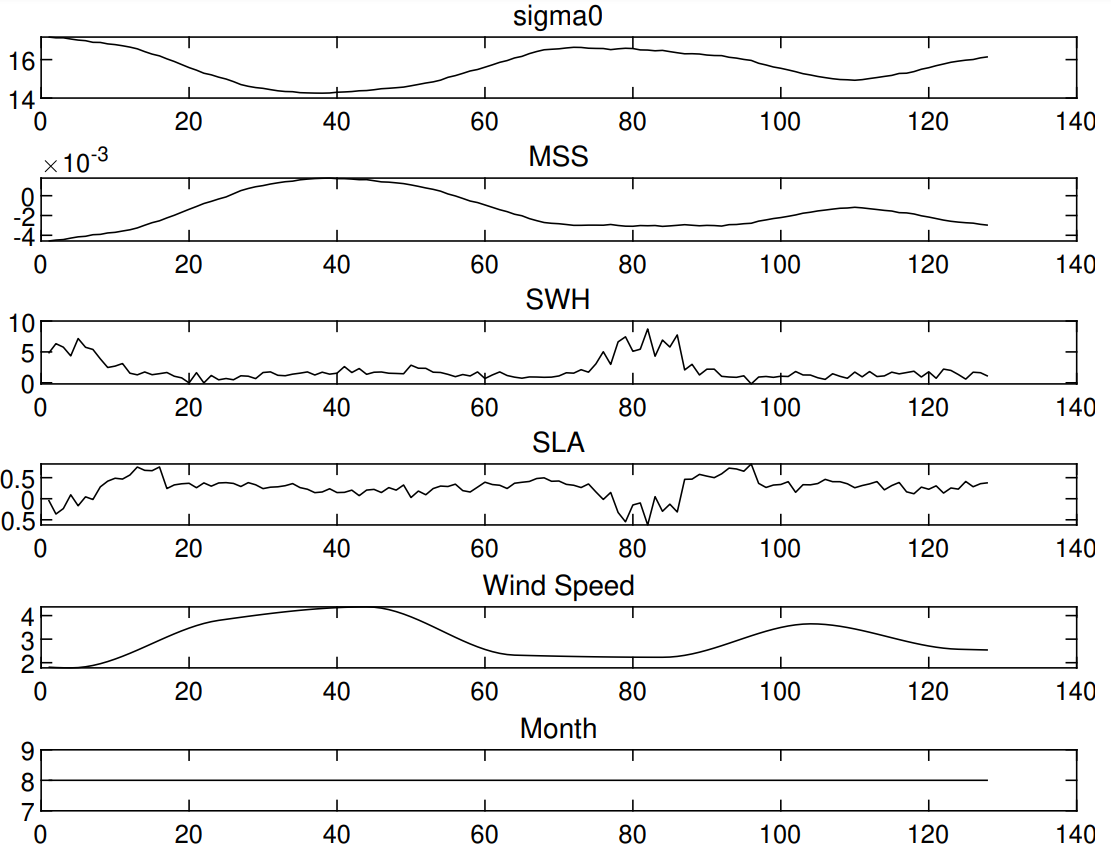}
\caption{\textbf{Data format of our dataset.} This example contains 6 features of 128 altimeter data points, i.e., a window sample containing spatially contiguous 128 data points. Each row corresponds to a feature vector along this window. For instance, the bottom row indicates the current month of the observation is August.}\label{fig_1}
\end{figure}

LRM satellite altimeter collects geophysical parameters on earth remotely and continuously, among which several parameters contain important information about ISWs. We thus utilize those parameters for ISW localization. Specifically, we choose the same ISW signature parameters as in \cite{bib8,bib9,bib10}, to be the data inputs of the ML models. They include radar backscatter (sigma0), differenced-mean-square slope (MSS), significant wave height (SWH), and sea level anomaly (SLA). We also take the month (time signature) and wind speed into consideration due to their correlation to ISW occurrence and its imaging mechanism \cite{bib62}. In summary, there are in total 6 features to describe whether an ISW exists in one location, as shown in Fig. \ref{fig_1}. In Fig. \ref{fig_1}, we visualize 128 data points that are spatially contiguous for demonstration. The horizontal axis indicates the spatial variance along satellite movement. The vertical axis is the value of each feature collected. As such, our original data is in the shape of (6,88576), where 6 is the number of features in a data point and 88576 is the total number of collected and labeled data points along the trajectory of satellites. For each column (which indicates a location, we refer to as a data point), we manually decide whether there is an ISW in this location by comprehensively considering the corresponding remote sensing images to maximize the labeling accuracy.

Furthermore, it's worth noting that the occurrence of an ISW in one location affects its neighborhood area, as explained in Section \ref{subsec_2_2} \cite{bib9,bib10}. To better locate ISWs, we make use of a window of data samples including not only the ISW data features of one current location (a data point) but also its spatially contiguous neighborhood. We choose to include 16 spatially contiguous altimeter observations as one data input given to an ML model, by directly slicing the dataset with a window size 16. Each data point in the window is represented by the 6 aforementioned features. There are 96 input features (6 times 16) in total in one sample for training. When using DNN algorithms (such as CNN), we give the 16 $\times$ 6 data matrix directly to the model as per data sample; when using traditional ML algorithms (such as Random Forest, LightGBM, and MLP), the data matrix is flattened into a vector with 96 feature values as per data sample. The choice of 16 observations guarantees that there is at most one ISW within the window according to the distance measurement of altimeter data. There are no overlapping samples between two adjacent windows. Thus, the original 88576 single-location data samples are transformed into 5536 window-based data inputs. The output of the model will be an integer between 0 and 16, telling which location in the window has an ISW. The output of 0 indicates no ISW in the window. By doing so, we formulate the ISW localization task into a 17-class classification problem. Due to the high cost of labeling by hand, our dataset has only 5536 (i.e., 88576/16) labeled samples in total. 

This multi-class classification task for fine-grained localization is defined below. We train and evaluate an ML model on our collected dataset $\mathcal{D}=\{(x_0,y_0),(x_1,y_1),(x_n,y_n),...,(x_N,y_N)\}$ where $N=5536$. Each sample $(x_i,y_i)$ contains input feature $x_i$ and output label $y_i$. The i-th input $x_i$ contains 96 input features from a window of 16 data points (single-location samples) with 6 altimeter features each. The output $y_i$ is in $\{0,1,2,3,...,16\}$, where $y_i=0$ indicates no ISWs in this window of sample while $y_i = k$ ($k \neq 0$) indicates ISW occurs on the $k^{th}$ data point of the sample. 

Before using the dataset for training, we perform the following pre-processing steps. First, we apply feature-wise standardization. For row-i (the i-th feature) in original data with shape (6,88576), mean values $\mu_i$ and standard deviations $\sigma_i$ are calculated. For the entry on the i-th row and j-th column of original data ($x_{ij}$), we minus the mean value of this feature and divide by its variance: $x_{ij}=(x_{ij}-\mu_i)/\sigma_i$. Then, we slice the standardized data into windows with a length of 16 as mentioned above. To accommodate different ML algorithms, each input sample in the dataset is transformed by either: 1) being flattened into a (1,96) vector for traditional ML algorithms, or 2) adding zero padding to augment the (6,16) matrix into a (16,16) matrix, as the input of some deep neural networks.

\subsection{ST-ECNN}
In this section, we first discuss the motivation for employing group equivariant networks and justify the intrinsic symmetries, i.e., scale and translation, in our defined task and dataset. Then, we introduce scale-translation group convolutions before outlining the proposed neural network architecture.

\subsubsection{Motivation}

To start with, the proposal of G-CNNs is motivated by: 1) loss of geometric guarantees (equivariance) in existing solutions \cite{bekkers2019b}, 2) symmetry that provides efficient representation learning through weight sharing \cite{bekkers2018roto,linmans2018sample}, and 3) the need of incorporating arbitrary symmetries \cite{cohen2021equivariant}. The sample efficiency and learning efficiency are outstanding properties by considering symmetries, which can well handle the aforementioned challenges in our scenario.

The symmetries in scale and translation are observed in our defined task which resembles a multi-variable multi-classification problem in time series. As a prior knowledge, ISW recognition heavily relies on the ``sudden'' \footnote{Note: depending on the scale of data, i.e., it could appear as an abrupt change if one compresses the data along the horizontal axis.} changes in local signals, leading to peaks and troughs of waves in visualizations. For instance, an ISW is presumed to be likely located around position 80 on the horizontal axis in Fig. \ref{fig_1}. As such, scaling the waves doesn't change the intrinsic properties, i.e., the shape of waves, which are significant for ISW localization. Likewise, translating the sample along the horizontal axis keeps the inherent features that represent an ISW as well. Both transformations applied to inputs will lead to predicted changes in outputs. As a result, scale and translation symmetries should be preserved in the learned mapping. Formally, we define the scale-translation group as follows.

\textbf{Scale-Translation Group $\mathcal{ST} = \mathbb{R}^2 \rtimes \mathbb{R}^+$.} The scale-translation group $\mathbb{R}^2 \times \mathbb{R}^+$ of translations vectors in $\mathbb{R}^2$ and scale factors in $\mathbb{R}^+$. It is equipped with the group product and group inverse:

\begin{align*} 
    g \cdot g' &= (\textbf{x}, s) \cdot (\textbf{x}', s') = (s\textbf{x}' + \textbf{x}, ss')\\
    g^{-1} &= (-\frac{1}{s}\textbf{x}, \frac{1}{s})
\end{align*}

where $g = (\textbf{x},s), g' = (\textbf{x}', s'), g \cdot g' = e = (\textbf{0}, 1)$.

\subsubsection{Group Equivariant Convolution}
We are to impose the equivariance to the scale-translation group by guaranteeing that the resulting features will shift accordingly if inputs are scaled or translated. To avoid overwhelming details and complex formulae in the theoretical modeling of equivariance using group theory, we refer readers to \cite{cohen2021equivariant} for more details, and only introduce the efficient implementations of group equivariant convolutions. For this purpose, we introduce a simpler and more intuitive perspective to understand group equivariance: transforming kernels instead of samples. One naive solution to equivariance is data augmentation. However, the augmentation is non-exhausted because there are endless elements in the, for instance, rotation group as images can arbitrarily rotate while maintaining their properties \footnote{Note: thinking about a malicious cancer image. It remains malicious despite the imaging angle. Refer to \cite{bekkers2018roto} for more details.}. Instead of transforming data for augmentation, one can transform convolutional kernels. This not only handles the issue of the infinite number of transformations but also achieves weight sharing (transformed patterns can also be detected by transformed kernels). In the following paragraphs, we shall introduce the three types of layers for group convolution: the lifting layer that lifts to group space, the group convolution layer that detects features through template matching, and the projection layer that projects back to $\mathbb{R}^d$ space.

\textbf{Lifting layer.} The lifting layer lifts the original data to group space such as $\mathcal{ST}$. In simple terms, we employ a set of kernels that discretize $\mathcal{ST}$ group. If it's a rotation group, one can discretize it into 16 kernels where each one rotates for i times ($i=0,1,...,15$) of $360/16=22.5^{\circ}$. The lifting layer is co-constituted by the set of kernels together. In our scenario, each kernel in this kernel set is a scale-translated version of the origin kernel. As compared to normal convolution, group convolution employs a set of transformed kernels rather than a set of transformed (or augmented) inputs. In this approach, when the input data is scale-translated, the learned kernels are still useful (weight sharing). In comparison with normal convolution, the output of the lifting layer contains an extra dimension that houses symmetry information (position and scale component). It is also worth noting that we can exploit the properties of the scale-translation group to simplify on top of the general formulation of group convolution \cite{romero2020wavelet}. As a result, the scale-translation lifting convolution can be regarded as a set of 1D convolutions with a series of scaled convolutional kernels \cite{romero2020wavelet}.

\textbf{Group convolution layer.} Once lifted to $\mathcal{ST}$ space, the group convolution is the same as the lifting layer. The kernel in the group convolution layer is a 3D kernel that assigns weights for each transformation. The convolution (or cross-correlation) is computed to match the templates all over the data. In this case, the feature detection of sequence signals is just the same as in normal convolution, i.e., template matching, except for an extra dimension for symmetry. Since there is a kernel to detect such kind of feature in each direction, imposing a scale and a translation just changes the position of the activated signal in the output feature map. Therefore, valuable model capacity is preserved by embedding geometry guarantees without losing any information. Similarly, the scale-translation convolution can also be simplified in terms of formulation, and this group convolution can be seen as a set of 1D convolutions with a series of scaled convolutional kernels followed by an integral over scales \cite{romero2020wavelet}. We refer readers to \cite{romero2020wavelet} for more rigorous formulations and definitions of group convolutions, including the general and scale-translation forms.

\textbf{Projection layer.} At the end of group convolution layers, the output feature map is shrunk down from group space to $\mathbb{R}^d$ space to make predictions. The final projection layer usually preserves the max or mean value over the extra component housing symmetry information to achieve invariance through max-pooling or mean-pooling. 

\begin{figure}[ht]%
\centering
\includegraphics[width=1.0\textwidth]{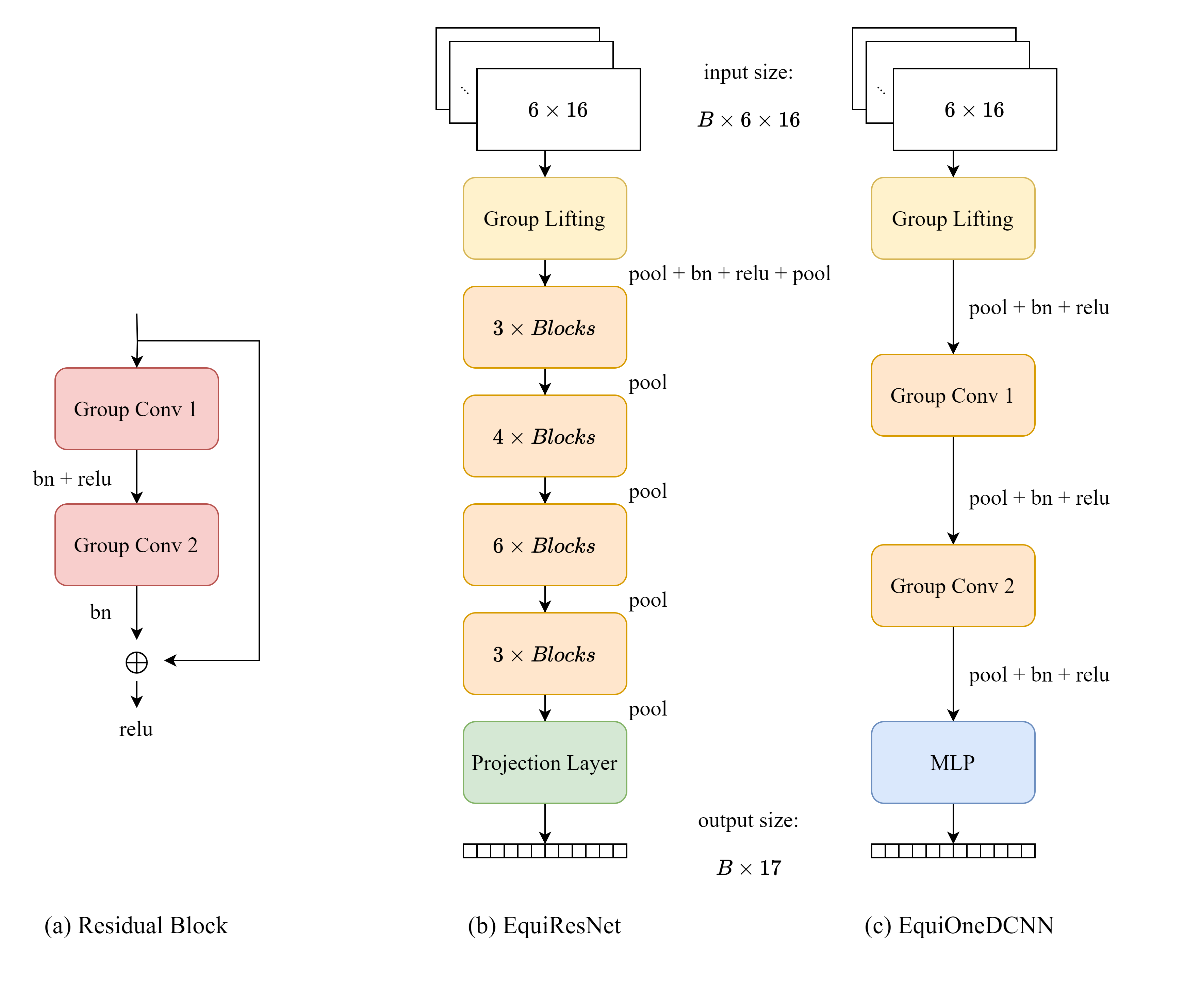}
\caption{\textbf{Architecture of EquiResNet and EquiOneDCNN.} (a) Residual block that constitutes EquiResNet. Group convolutions are accompanied by batch normalization and ReLU activation. (b) EquiResNet contains a group lifting module that first lifts to $\mathcal{ST}$ space, followed by a series of residual blocks. The final projection layer is a group convolution and max pooling across the scale axis, which projects back to $\mathbb{R}$ space. (c) EquiOneDCNN contains a lifting layer followed by two group convolution modules. The pooling operation before prediction head MLP serves as the projection module to downgrade back to $\mathbb{R}$ space.}\label{fig:archi}
\end{figure}

\subsubsection{Architecture}

For ST-ECNN, we propose two types of architectures for experiments. Following ResNet, we introduce EquiResNet as shown in Figure \ref{fig:archi}. It first lifts to $\mathcal{ST}$ space using lifting layers discussed in the previous section. Then, an arbitrary number of group convolution layers are stacked to conduct template matching (feature detection) on group space. Here, we use the same architecture as ResNet-50, namely, [3, 4, 6, 3] residual blocks for each stage. For each block, two group convolutions are utilized in combination with element-wise addition as the residual connection. Finally, a max-pooling layer serves as invariant mapping to project signals back to $\mathbb{R}$ since output space should accord with input space.

We further propose a more lightweight architecture for comparison. EquiOneDCNN contains a simple lifting layer followed by two group convolution layers. The MLP module is appended at the bottom for predictions. It's worth mentioning that the pooling operation between the second group convolution and MLP has projects back to $\mathbb{R}$ space. Thus, no independent projection layer is added.

\subsection{Evaluation} \label{subsec_3_2}

Regarding performance evaluation in the class imbalance learning area, geometric mean (G-mean), F1-score, area-under-curve of receiver-operating-characteristic (ROC-AUC), and Matthews correlation coefficient (MCC) are more commonly used than vanilla accuracy metrics, because a model can achieve high accuracy even though its performance in the minority class is very poor. Beyond binary classification, multi-class AUC (MAUC) and multi-class MCC (MMCC) are widely employed for multi-class classification evaluation. In this work, we choose to use G-mean, MAUC, and MMCC to evaluate our algorithms under imbalanced scenarios based on \cite{bib20}. 
G-mean is defined as the geometric mean of recall of all classes \cite{bib64}. It is the most popular ``overall" performance metric, which can reflect the accuracy of all classes. 

AUC indicates the model's capacity for distinguishing between positive and negative samples in binary classification. It is often discussed in class imbalance studies because of its insensitivity to the imbalance ratio of data. MAUC is its extension for multi-class data \cite{bib18}, which is defined as below: 

\begin{equation}\label{eq_mauc}
MAUC = \frac{2}{C(C-1)}\sum_{i<j}A(i,j)
\end{equation}

where in Eq. \ref{eq_mauc}, C is the number of classes, and A(i,j) denotes AUC between class i and j. 

MCC is commonly used as a metric for imbalanced biomedical data \cite{bib20}, while MMCC serves as a generalization of MCC to multi-class scenarios \cite{bib19,bib21}. MMCC is formally defined as below \cite{bib23}:

\begin{equation}
MMCC = \frac{c*s-\sum_{k}^{K}p_k*t_k}{\sqrt{(s^2-\sum_{k}^{K}{p^2_k})*(s^2-\sum_{k}^{K}{t^2_k})}}
\end{equation}

\noindent
where K is the number of classes, and $C_{ij}$ denotes the observation number of class-i predicted to be class-j in the confusion matrix.

In addition,

\begin{itemize}
\item $t_k = \sum_{i=1}^{K}C_{ik}$ the number of class-k samples (with true label k),
\item $p_k = \sum_{i=1}^{K}C_{ki}$ the number of samples predicted as class k,
\item $c = \sum_{k=1}^{K}C_{kk}$ the total number of correctly predicted samples,
\item $s = \sum_{i=1}^{K}\sum_{j=1}^{K}C_{ij}$ the total number of samples.
\end{itemize}

In ISW localization, there exists an issue when evaluating the accuracy of a model -- how accurately predicting a location should be counted as a correct classification? The aforementioned metrics provide a hard line between correctness and incorrectness. When predicting the location of an ISW, however, such hard separation is not always appropriate. If the exact location of an ISW data sample is at location number 10, the question is: whether a prediction giving location 9 is wrong. Domain experts accept some level of variations in locations. Therefore, we propose k-approximate accuracy to measure classification accuracy with tolerance degree k. A prediction is treated as correct if the distance between the predicted location and the true location is no larger than k. It allows a tolerable range in location prediction instead of the exact single-point prediction.

k-approximate accuracy is calculated based on the following definition: For the case with an ISW (i.e. the ground truth $y \neq 0$), if the prediction of the model $\hat{y}$ is no more than k data points far from $y$ (i.e. $\mid \hat{y}-y \mid \leq k$), the prediction is treated as correct; for the case without an ISW ($y = 0$), the correct prediction $\hat{y}$ must be 0. In this work, we set k to 1, 3, and 5 based on domain knowledge, allowing at most 5 location differences in prediction. In the following experiment tables, we will term it as ``Acc-k".

\section{Experiments}\label{sec_4}

In this section, we empirically verify the performance of our proposed methods. We first compare ST-ECNNs with recently developed strong CNN baselines. Then, an empirical study is conducted to probe self-supervised pre-training in our scenario.

\subsection{Experiment 1: Comparisons between different methods}\label{subsec_4_1}

We start by considering different baseline methods including Random Forest (RF) \cite{breiman2001random}, LightGBM (LGB) \cite{ke2017lightgbm}, vanilla Multi-Layer Perceptron (MLP), ResNet \cite{bib1}, BoTNet \cite{bib12}, and ConvNeXt \cite{liu2022convnet}. It's worth noting that the input sample has been adaptively reshaped to cater to different methods. For ST-ECNN, we mainly compare two architectures described in the previous section: EquiOneDCNN and EquiResNet.

\subsubsection{Experimental Settings}\label{subsubsec_4_1_1}

In our following experiments, we adopt 5 $\times$ 2-fold cross-validation as the evaluation strategy to better test generalization capacity. Different hyper-parameter settings are considered and tested for each candidate algorithm, among which the one achieving the best performance joins the model comparison. For more implementation details, one is referred to our code repository here: \url{https://github.com/ZhangWan-byte/Internal_Solitary_Wave_Localization}.

For Random Forest and LightGBM, the total number of base trees and the maximum depth of each tree are set to 10. For MLP, we fine-tune the number of layers and neurons, leading to a model with 4 hidden layers and the numbers of neurons of [512,256,128,64]. Each layer consists of a linear transformation, ReLU activation, and batch normalization \cite{bib43}. The loss function is cross entropy. The batch size is set to 1024. We use Adam \cite{bib42} as the optimizer with learning rate of 3e-4, and train the model for 200 epochs. 

\begin{table}[htp]
\centering
\caption{The mean and standard deviation of the eight ML algorithms on the six performance metrics.}
\label{table_1}
\renewcommand{\arraystretch}{1.5} 
\resizebox{\textwidth}{!}{%
\begin{tabular}{@{}ccccccc@{}}
\toprule
         & G-Mean            & MAUC              & MMCC              & Acc-1       & Acc-3             & Acc-5             \\ \midrule
RF       & 0.383±0.033       & 0.734±0.013       & 0.333±0.048       & 0.818±0.011 & 0.822±0.012       & 0.823±0.012       \\
LGB      & 0.486±0.040       & 0.746±0.014       & 0.448±0.050       & 0.836±0.014 & 0.839±0.013       & 0.839±0.013       \\ \midrule
ResNet   & {\ul 0.696±0.036} & {\ul 0.926±0.010} & {\ul 0.620±0.033} & 0.901±0.010 & 0.924±0.007       & 0.926±0.008       \\
BoTNet   & 0.690±0.024       & 0.922±0.009       & 0.609±0.025       & 0.896±0.009 & 0.921±0.006       & 0.923±0.006       \\
ConvNeXt & 0.659±0.007       & 0.830±0.013       & 0.566±0.020       & 0.857±0.013 & 0.889±0.011       & 0.897±0.010       \\ \midrule
OneDCNN  & 0.655±0.010       & 0.894±0.009       & 0.612±0.008       & 0.907±0.006 & {\ul 0.935±0.004} & {\ul 0.937±0.003} \\
EquiOneDCNN & 0.639±0.017          & 0.919±0.007          & 0.615±0.020          & \textbf{0.922±0.008} & \textbf{0.945±0.003} & \textbf{0.946±0.004} \\
EquiResNet  & \textbf{0.721±0.053} & \textbf{0.945±0.022} & \textbf{0.653±0.049} & {\ul 0.920±0.015}    & {\ul 0.935±0.010}    & {\ul 0.937±0.009}    \\ \bottomrule
\end{tabular}%
}
\end{table}

We adopt the standard ResNet-50 in the experiment as discussed in \cite{bib1}. For a fair comparison, the BoTNet-50 version of BoTNet joins the experiment, as the counterpart architecture of ResNet-50. The only difference between BoTNet-50 from ResNet-50 is the replacement of the convolutional layer with the MHSA layer in stage 5 of ResNet \cite{bib12}. For both ResNet-50 and BoTNet-50, the batch size is set to 1024; the Adam optimizer with the learning rate of 3e-4 is employed; the training epoch is set to 400. 

To verify the performance of equivariant networks, we further add a non-equivariant counterpart of EquiOneDCNN, i.e., OneDCNN, for comparison. It shares the same architecture as EquiOneDCNN except that all of its modules are not equivariant. The architecture details of EquiOneDCNN and EquiResNet are elaborated in the previous section.

Under the above settings, there are eight models trained and compared. All experiments are run on a Windows laptop with Intel(R) Core(TM) i7-11800H CPU @ 2.30GHz, RAM 32 GB. The GPU is an NVIDIA GeForce RTX 3070 Laptop. The average value and the standard deviation of G-mean, MAUC, MMCC, and k-approximate accuracy (Acc-k, k = 1, 3, 5) are recorded and reported. The Mann-Whitney U test at 95\% confidence level is adopted to examine the significance of performance difference.

\begin{figure}[ht]%
\centering
\includegraphics[width=1.0\textwidth]{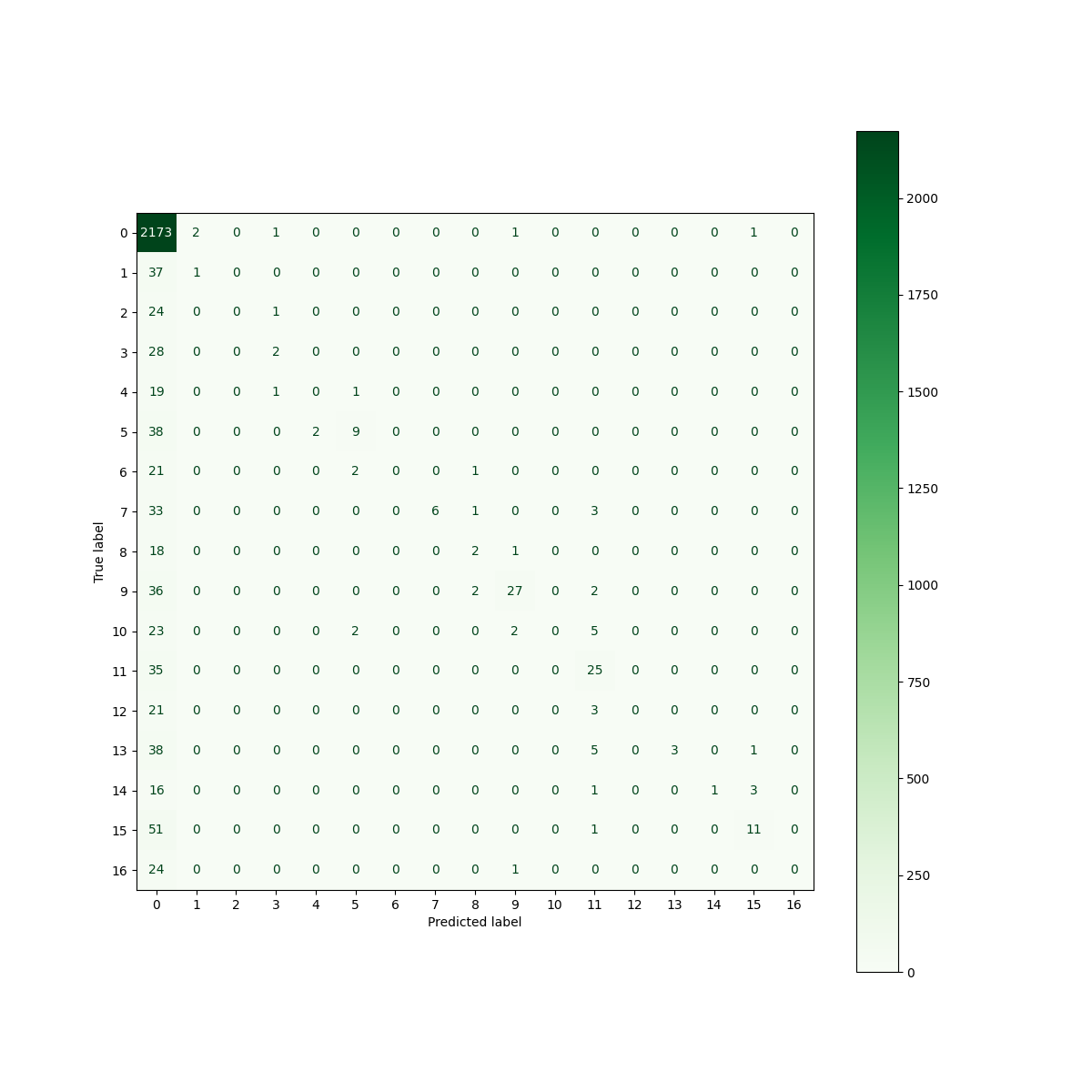}
\caption{RF Confusion Matrix}\label{fig_2}
\end{figure}

\subsubsection{Result Analysis}\label{subsubsec_4_1_2}

The experimental results are shown in Table \ref{table_1}, where the best scores are bolded and the second bests are underlined. We observe that EquiResNet achieves the highest scores in terms of G-Mean, MAUC, and MMCC, on which normal ResNet achieves the second-highest scores. Furthermore, EquiResNet maintains the second on k-accuracy metrics, only below EquiOneDCNN. It is observed that EquiResNet and EquiOneDCNN are good at imbalanced metrics and approximate accuracy, respectively. We speculate that it is because EquiResNet, as a stronger learner due to its larger capacity, could capture subtle patterns reflected by imbalanced metrics. In comparison, EquiOneDCNN has a more concise assumption due to its lower capacity. Therefore, it achieves less accurate (misclassified as contiguous classes) and acceptable performance (on imbalanced metrics). We empirically find that variants of ResNet such as BoTNet and ConvNeXt show no advantage on our task and dataset compared to vanilla ResNet. We believe it is due to a different scenario from computer vision tasks. As such, architectures specifically designed for imaging scenarios are less effective in our problem formulation. Traditional machine learning algorithms including Random Forest and LightGBM suffer from poor performance due to constrained model capacity.

\begin{figure}[ht]%
\centering
\includegraphics[width=1.0\textwidth]{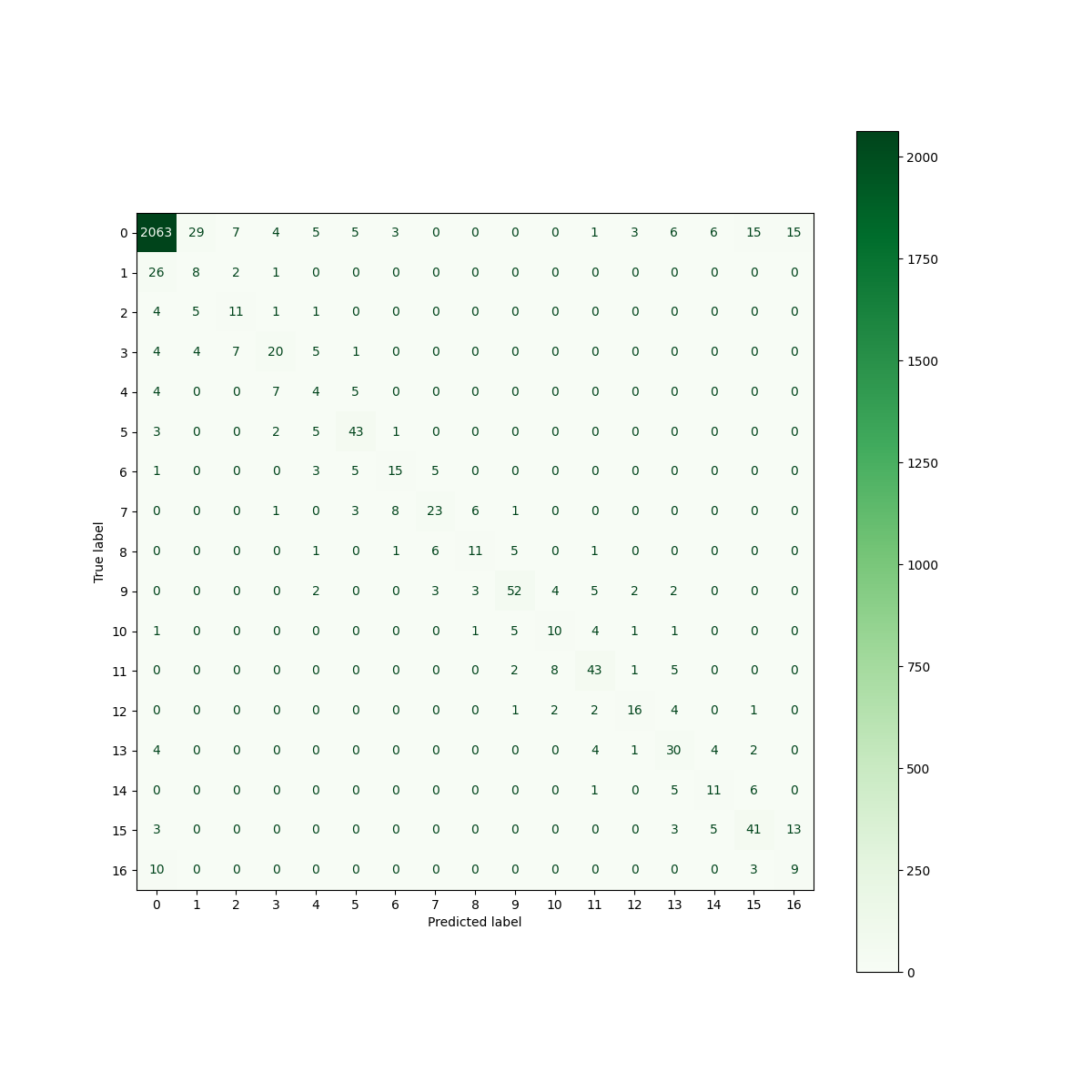}
\caption{EquiResNet Confusion Matrix}\label{fig_3}
\end{figure}

To further understand which classes cause the performance difference among the models, especially between the worst model RF and the best model EquiResNet, we present their 17$\times$17 confusion matrices in Fig. \ref{fig_2} and Fig. \ref{fig_3}. Please note that we use positive/negative in the following analysis to describe instances with/without ISWs. The key difference lies in the first column, indicating that more positive instances are misclassified as negative by RF. In contrast, EquiResNet has a less false negative rate (the entries in the first column) than RF and a higher accuracy of overall prediction. However, RF demonstrates a higher true-negative (TN) rate with 2163 TN samples, while EquiResNet only correctly predicts 2063 TN samples. We speculate that it is because EquiResNet, as a stronger ML model, is more capable of capturing minority class features, and thus tends to recognize the existence signatures of ISWs, leading to an inclination of positive prediction. In comparison, due to the limited model capacity, RF can only extract lower-level features, and therefore make conservative predictions, tending to classify an input sample as negative.

We also observe that the location of an ISW is frequently predicted to be the adjacent locations to ground truth. This is because different ISWs have different sizes, and an ISW can span several locations. Thus, it is technically not wrong to predict a data point positive if it's contiguous to our labeling position. However, because the dataset only labels single-point position as the ground truth, the spatially contiguous property of the localization problem is ignored. This is the reason we propose and discuss k-approximate accuracy. 

\subsection{Experiment Part 2: Exploiting Massive Unsupervised Data with Contrastive Pre-Training}\label{subsec_4_2}

Self-supervised learning provides a powerful tool to utilize the massive unlabeled data, which we believe could complement the challenge of the limited quantity of supervised data. Therefore, in this section, we are motivated to make use of the knowledge from a large amount of unlabeled data by employing contrastive learning (CL) as reviewed in Section \ref{subsec_2_4}. Through pre-training on unlabeled data, prior knowledge can also be introduced into the model before learning from the labeled data. We expect that this can improve the generalization ability of EquiResNet.

\subsubsection{Experimental Settings}\label{subsubsec_4_3_1}
Contrastive pre-training is conducted on a large amount of unlabeled altimeter data with 69910 samples from the ocean regions of the South China Sea, Andaman Sea, and Sulawesi Sea. We expect to employ a larger unsupervised dataset given more available computational resources. An encoder model is trained contrastively under the SimCLR framework as described in Section \ref{subsec_2_4}. An ideal encoder projects the original features to another vector space where the distance between similar instances (samples of the same class) is smaller than the distance between dissimilar instances (samples of different classes). We then use the encoder's pre-trained weights to initialize EquiResNet for supervised training. Finally, we follow the same training process as in Section \ref{subsec_4_1} to train and evaluate the model. In the next section, we first explore the adaptability of CL on our ISW dataset using ResNet. Then, we compare the fine-tuned ResNet and EquiResNet models.

\subsubsection{Result Analysis}\label{subsubsec_4_3_2}

According to \cite{bib27}, SimCLR benefits from a large batch size and properly adjusted temperature coefficient $\tau$. As discussed in Section \ref{subsec_2_4}, this is because a large batch size brings more pair-wise training samples, and the temperature coefficient is vital to the degree of clustering. To find the best batch and $\tau$ settings for our CL pre-training, we compare the ResNet models pre-trained with the batch size varying in [128,256,512,1024] and $\tau$ varying in [0.01,0.1,10,100]. The pre-trained weights are used to initialize another ResNet which is trained in supervise with cross-entropy loss with batch size 128, learning rate 3e-4, and epoch 400. For augmentation techniques, we utilize random resize cropping and random horizontal flipping. More augmentation techniques are to be explored in future works.

\begin{table}[htp]
\centering
\caption{Model performance of ResNet in different batch sizes and $\tau$ values}
\label{table_3}
\renewcommand{\arraystretch}{1.5} 
\resizebox{\textwidth}{!}{%
\begin{tabular}{llllll}
\hline
$\tau$               & batch size & 128         & 256                  & 512                  & 1024                 \\ \hline
\multirow{6}{*}{0.1} & G-mean    & 0.676±0.043 & 0.694±0.030          & \textbf{0.735±0.019} & 0.720±0.016          \\
                     & MAUC       & 0.906±0.025 & 0.919±0.017          & \textbf{0.933±0.013} & 0.923±0.012          \\
                     & MMCC       & 0.608±0.036 & 0.626±0.025          & \textbf{0.668±0.022} & 0.654±0.020          \\
                     & Acc-1     & 0.898±0.014 & 0.906±0.009          & \textbf{0.915±0.007} & 0.909±0.010          \\
                     & Acc-3     & 0.920±0.009 & 0.928±0.006          & \textbf{0.934±0.006} & 0.931±0.007          \\
                     & Acc-5     & 0.922±0.009 & 0.930±0.006          & \textbf{0.936±0.006} & 0.932±0.007          \\ \hline
batch                & $\tau$     & 0.01        & 0.1                  & 10                   & 100                  \\ \hline
\multirow{6}{*}{512} & G-mean    & 0.723±0.024 & \textbf{0.735±0.019} & 0.678±0.027          & 0.726±0.021          \\
                     & MAUC       & 0.915±0.017 & \textbf{0.933±0.013} & 0.918±0.015          & 0.937±0.010          \\
                     & MMCC       & 0.666±0.018 & \textbf{0.668±0.022} & 0.616±0.029          & 0.655±0.027          \\
                     & Acc-1     & 0.909±0.008 & \textbf{0.915±0.007} & 0.907±0.010          & 0.915±0.010          \\
                     & Acc-3     & 0.933±0.006 & \textbf{0.934±0.006} & 0.927±0.007          & 0.931±0.009          \\
                     & Acc-5     & 0.934±0.005 & \textbf{0.936±0.006} & 0.928±0.007          & 0.932±0.008          \\ \hline
\multirow{6}{*}{1024} & G-mean & 0.714±0.015 & 0.720±0.016 & 0.731±0.024 & \textbf{0.762±0.025} \\
                     & MAUC       & 0.919±0.009 & 0.923±0.012          & 0.935±0.011          & \textbf{0.955±0.009} \\
                     & MMCC       & 0.651±0.016 & 0.654±0.020          & 0.674±0.018          & \textbf{0.695±0.021} \\
                     & Acc-1     & 0.904±0.006 & 0.909±0.010          & 0.920±0.007          & \textbf{0.930±0.006} \\
                     & Acc-3     & 0.930±0.006 & 0.931±0.007          & 0.937±0.005          & \textbf{0.940±0.003} \\
                     & Acc-5     & 0.931±0.006 & 0.932±0.007          & 0.937±0.004          & \textbf{0.940±0.004} \\ \hline
\end{tabular}
}
\end{table}

Key results are shown in Table \ref{table_3} \footnote{Note that experiments in this table are conducted using a different device due to the large computational overhead, which may introduce other factors that cause unexplainability. ResNet (instead of EquiResNet) is also used for the same reason. Thus, resulting values in this table should not be directly compared with those in other tables.}. The first two rows show results of different batch sizes when $\tau=0.1$. The next three rows show results of different $\tau$ values when batch size is 512 and 1024, where the best two results are bolded. As expected, although ResNet with batch size 512 achieves the best over the other batch size settings, the largest batch size provides the best final overall results. We further discover that a larger $\tau$ brings a similar impact. In the projected space, a large $\tau$ tends to form the samples of the same class into clusters. As Eq. \ref{eq_2} indicates, to minimize the loss, $sim(z_i,z_j)$ (distance between positive pairs) is maximised and $sim(z_i,z_k)$ (distance between negative pairs) is minimised. In other words, such a training process requires the encoder to project original data into a space where similar instances are near while dissimilar instances are distant. A larger $\tau$ reduces the distance of all pairs, making the distance distribution milder, and therefore forces the encoder to further maximize the similarity between positive pairs. In our scenario, forming different samples into clusters is without doubt beneficial to our task. The differences between samples from the same class are not so important in our case, as they're highly similar. Therefore, we would like the clustering effect to be as obvious as possible, suggesting a larger $\tau$. To step further, a smaller batch size provides a more narrow range of negative samples with less diversity. Therefore, a smaller batch size makes it harder to distinguish between positive and negative samples during pre-training, which precisely calls for a smaller temperature factor $\tau$ since this sharpens the distribution and thus makes learning easier.

\begin{table}[ht]
\centering
\caption{Results of pre-trained \& fine-tuned EquiResNet as compared to other equivariant networks.}
\label{table_4}
\renewcommand{\arraystretch}{1.5} 
\resizebox{\textwidth}{!}{%
\begin{tabular}{ccccccc}
\hline
             & G-Mean            & MAUC              & MMCC              & Acc-1             & Acc-3                & Acc-5                \\ \hline
EquiOneDCNN  & 0.639±0.017       & 0.919±0.007       & 0.615±0.020       & {\ul 0.922±0.008} & \textbf{0.945±0.003} & \textbf{0.946±0.004} \\
EquiResNet   & {\ul 0.721±0.053} & {\ul 0.945±0.022} & {\ul 0.653±0.049} & 0.920±0.015       & 0.935±0.010          & 0.937±0.009          \\
EquiResNet+ & \textbf{0.726±0.038} & \textbf{0.949±0.009} & \textbf{0.668±0.032} & \textbf{0.930±0.003} & {\ul 0.943±0.003} & {\ul 0.944±0.003} \\
EquiResNet++ & 0.706±0.033       & 0.924±0.013       & 0.628±0.030       & 0.914±0.008       & 0.932±0.006          & 0.935±0.006          \\ \hline
\end{tabular}%
}
\end{table}

Based on the observations as discussed above, we report contrastive pre-trained results of EquiResNet in Table \ref{table_4}. It's worth noting that EquiResNet+ is pre-trained with batch size 128 and $\tau$ 0.1 while EquiResNet++ is pre-trained with batch size 128 and $\tau$ 10. We anticipate conducting more fine-grained hyper-parameter tuning (and larger batch size) given abundant computational resources in the future. It is observed in Table \ref{table_4} that, for EquiResNet, small $\tau$ (0.1) with a small batch size (128) achieves better results than using $\tau=10$, which further verifies our previous analyses. Despite a relatively smaller pre-training batch size, pre-trained EquiResNet also achieves an overall better performance than that without pre-training.

\section{Conclusion}\label{sec_5}

\textbf{Discussion.} In this paper, we propose a novel approach to detect and locate ISWs at the finest level of grain. We first justify the advantages of locating ISWs based on altimetry instead of imagery, and handcraft a new dataset from scratch using satellite Jason-2/3 altimetry data. Then, we formulate the localization problem as an intra-window multi-class classification problem based on the intrinsic pattern of ISWs. With our task formulation, the challenges mainly lie in the limited quantity of labeled data, which we propose could be alleviated through prior knowledge injection. Therefore, we exploit the inherent symmetries of scale and translation and introduce the ST-ECNN with two specific architectures: EquiOneDCNN and EquiResNet. Then, we propose to inject prior knowledge by pre-training on massive unsupervised data using the SimCLR framework. A rigorous comparison based on the Mann-Whitney U test at 95\% confidence is made through 5 times 2-fold cross-validation. Approximate accuracy metrics are proposed for complementary evaluation beyond commonly used imbalanced metrics. The empirical results show that our proposed method achieves an overall better performance than baseline models. We believe our solution has comprehensively tackled the common and primal challenge in remote sensing scenarios. For future work, we look forward to further extending the principle of prior knowledge injection and developing better methods under this core motivation.

\textbf{Limitations.} Our existing solution mainly presents two limitations: 1) high computational complexity of group equivariant convolution, and 2) less effective means to deal with imbalanced data. For one thing, we anticipate more efforts devoted to developing computation-friendly group convolutions and better methods for imbalanced learning. For another thing, we aim to better incorporate prior knowledge for remote sensing tasks. Specifically, one potentially feasible method in our scenario is considering ``hard negatives'' \cite{robinson2020contrastive} during pre-training, because existing vanilla SimCLR doesn't consider intra-class relations (as prior knowledge) but blindly regards all other classes as negative samples. We sincerely hope more novel ideas to achieve efficiency from the view of prior knowledge incorporation could be proposed or introduced in the future.

\vskip 0.2in
\bibliography{main.bib}

\end{document}